%% file: main.tex
\documentclass[10pt,twocolumn,letterpaper]{article}

\usepackage{iccv}              % To produce the CAMERA-READY version
\input{preamble}

\definecolor{iccvblue}{rgb}{0.21,0.49,0.74}
\usepackage[pagebackref,breaklinks,colorlinks,allcolors=iccvblue]{hyperref}

\def\paperID{11907} % *** Enter the Paper ID here
\def\confName{ICCV}
\def\confYear{2025}

\title{Improving Multimodal Learning via Imbalanced Learning}

\author{Shicai Wei\textsuperscript{1,2} \quad\quad\quad Chunbo Luo\textsuperscript{1}* \quad\quad\quad  Yang Luo\textsuperscript{1}  \\
\textsuperscript{1}University of Electronic Science and Technology of China \\
\textsuperscript{2} National and Local Joint Engineering Research Center for Cloud Operating System \\
{\tt\small shicaiwei@std.uestc.edu.cn,  \{c.luo, luoyang\}@uestc.edu.cn }
}
\begin{document}
\maketitle

\input{sec/0_abstract}

% \input{sec/1_intro}
% \input{sec/2_formatting}
% \input{sec/3_finalcopy}
{
    \small
    \bibliographystyle{ieeenat_fullname}
    \bibliography{main}
}

\end{document}

%% file: preamble.tex
\usepackage[dvipsnames]{xcolor}

\usepackage{diagbox}
\usepackage{cases}
\usepackage{algorithmic}
\usepackage{algorithm}
\usepackage{array}
\usepackage{multirow}

%% file: sec/0_abstract.tex
\begin{abstract}

Multimodal learning often encounters the under-optimized problem and may perform worse than unimodal learning. Existing approaches attribute this issue to imbalanced learning across modalities and tend to address it through gradient balancing. However, this paper argues that balanced learning is not the optimal setting for multimodal learning. With bias-variance analysis, we prove that imbalanced dependency on each modality obeying the inverse ratio of their variances contributes to optimal performance. To this end, we propose the Asymmetric Representation Learning(ARL) strategy to assist multimodal learning via imbalanced optimization. ARL introduces auxiliary regularizers for each modality encoder to calculate their prediction variance.  ARL then calculates coefficients via the unimodal variance to re-weight the optimization of each modality, forcing the modality dependence ratio to be inversely proportional to the modality variance ratio. Moreover, to minimize the generalization error, ARL further introduces the prediction bias of each modality and jointly optimizes them with multimodal loss. Notably, all auxiliary regularizers share parameters with the multimodal model and rely only on the modality representation. Thus the proposed ARL strategy introduces no extra parameters and is independent of the structures and fusion methods of the multimodal model. Finally, extensive experiments on various datasets validate the effectiveness and versatility of ARL.  Code is available at  \href{https://github.com/shicaiwei123/ICCV2025-ARL}{https://github.com/shicaiwei123/ICCV2025-ARL}

\end{abstract}

\section{Introduction}
\label{sec:intro}

Multimodal learning has gained considerable attention across a wide range of applications, including action recognition~\cite{ar1,ar3}, object detection~\cite{mm_detection2,mm_detection3}, and audiovisual speech recognition~\cite{av1,av2}. While multimodal learning has significant potentiality in improving the robustness and performance of models, the heterogeneity of multimodal data raises the challenges to leverage multimodal correlations and complementarities.  According to recent research~\cite{ogm,umt,pmr}, the performance of existing multimodal methods is sub-optimal since the performance of each modality encoder falls significantly short of their unimodal bound. 

% They attribute this under-optimized phenomenon to the "modality imbalance" problem, where the dominance of a particular modality impedes the comprehensive exploitation of the multimodal data.

\begin{figure*}[ht]
\centering
\includegraphics[width=1.0\textwidth]{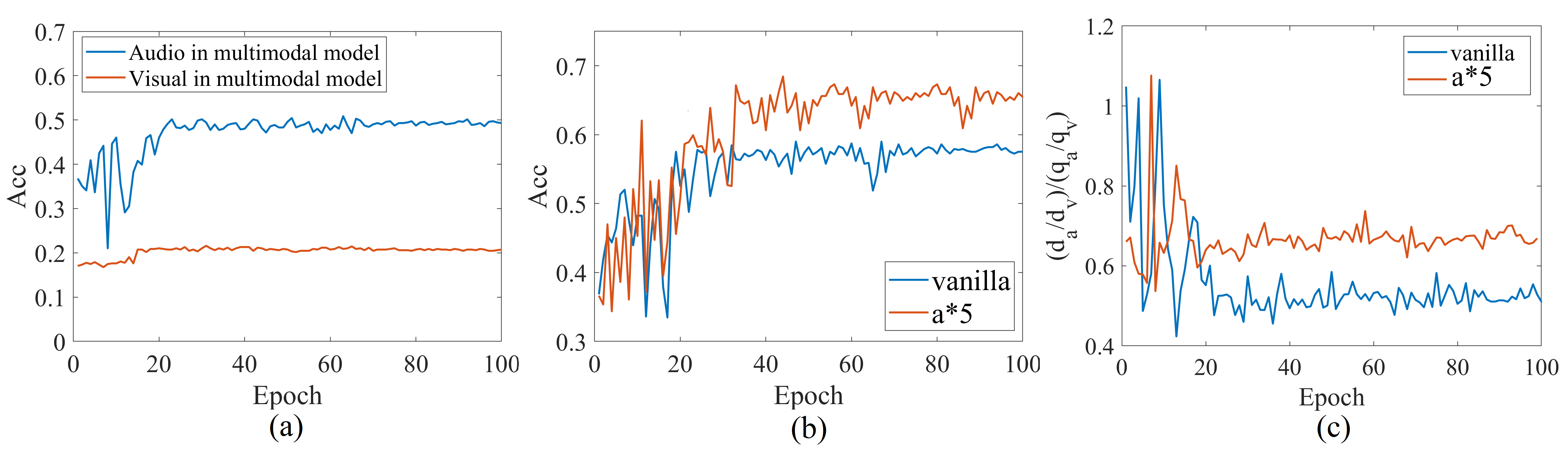} % Reduce the figure size so that it is slightly narrower than the column. Don't use precise values for figure width.This setup will avoid overfull boxes.
\caption{ The visualization on the audio-visual dataset CREMA-D. (a) presents the performance of the unimodal branch in the vanilla multimodal model. (b) presents the performance of the multimodal model with audio branch modulation. `$a*5$' represents increasing the gradient for the audio branch by fivefolds to accelerate its optimization. (c) presents the gap between the modality dependence and modality variance. $\frac{d_a}{d_v}$ denotes the model optimization dependence ratio on the audio and visual modalities (see Equation~\ref{impact}). $\frac{q_a}{q_v}$ denotes the inverse of the variance ratio of audio and visual modalities (see Equation~\ref{qq}).}
\label{diveristy-compare}
% \vspace{-0.5em}
\end{figure*}

Existing methods attribute this under-optimized phenomenon to the "modality imbalance" problem, where the dominant modality suppresses the learning of weak modality, hindering the full utilization of multimodal data~\cite{wh,ogm,umt,pmr,agm,MLA}. They then propose several methods to address the problem. Some of them aid the training of the weak modality with the help of additional unimodal classifiers~\cite{wh,mmpareto}, pre-trained models~\cite{umt} or modality resampling~\cite{FMV}. Since they will introduce additional neural modules and complicate the training procedure, other methods~\cite{ogm,pmr,agm} try to modulate the gradient of different modalities within the multimodal model.  Moreover, recent methods transform the conventional joint multimodal learning process into an alternating unimodal learning process to minimize inter-modality interference directly~\cite{MLA,reconboost,diagnosing}. 

% Specifically, they mine the modality that dominates the multimodal optimization and then balance its learning with the weak modalities by balancing their gradients.
% via the unimodal performance in the multimodal model

% While existing gradient-modulated methods achieve good performance, they still have some limitations. Firstly, our experimental findings indicate that gradient balancing is not insufficient for the model to balance the imbalanced optimization. Take the CREMA-D dataset as an example, the bigger modulation coefficient for under-optimized visual modality could contribute to higher performance. Secondly, all of the modality encoders are under-optimized compared to unimodal learning, including the dominant modality. More importantly, the gradient modulation for the weaker modality does not always contribute to better performance. In contrast, the key is to improve the performance-dominant modality, even if it is the optimization-domination modality.

While existing methods achieve good performance, they simply attribute the inferior multimodal performance to the imbalanced learning between modalities. This paper argues that this assertion may be imperfect. According to the assertion, for the CREMA-D datasets where the audio branch outperforms the visual branch (see Fig.~\ref{diveristy-compare} (a)), increasing the gradient of the audio branch should exacerbate the imbalanced learning and decrease performance. However, when we increase the gradient of the audio branch by 5 times, the model performance does not decrease but increases (see Fig.~\ref{diveristy-compare} (b)).  This finding drives us to further explore the real bottleneck that limits multimodal learning performance.

To this end, we reframe the conventional joint multimodal learning process by transforming it into the ensembling learning of multiple unimodal processes. From the perspective of bias-variance, we prove that the optimal decision dependence ratio on each modality should obey the inverse of their prediction variance. And conventional balanced learning is optimal only when the variances of the different modalities are the same. Therefore, instead of balancing multimodal learning, we should modulate the gradient of each modality to adjust their contribution ratio to the model optimization, enabling it to satisfy the inverse of the modality variance ratio. As shown in Fig.~\ref{diveristy-compare} (c), increasing the gradient of the audio branch in the multimodal model exactly alleviates the gap between the optimization dependence and the inverse of modality variance ratio.

To this end, we propose the Asymmetric Representation Learning (ARL) strategy to assist multimodal learning. ARL first introduces auxiliary regularizers for each modality encoder, which measure their variance based on the prediction logit output. Then ARL introduces asymmetric coefficients to modulate the gradients of each modality to adjust the optimization dependency on each modality, aligning it with the inverse of the modality variance ratio. Additionally, ARL incorporates the original gradient as a residual component to the adjusted gradient, ensuring that encoder optimization continues even when the modulation factor is extremely small. Furthermore, to reduce generalization error, ARL integrates the prediction bias of each modality and optimizes them in conjunction with the multimodal loss. Importantly, all auxiliary regularizers share parameters with the multimodal model and depend solely on the modality representations. Thus, the ARL strategy introduces no extra parameters and is agnostic to the architecture and fusion techniques of the multimodal model. We conduct extensive experiments in various multimodal tasks on different datasets to study the effectiveness and versatility of ARL. The contributions are summarized as follows,

\begin{itemize}
    % \item We reveal the under-optimized problem of each modality in the multimodal model, including the modality with higher performance, caused by the gradient coupling.
    \item  We reveal that the balanced learning of different modalities is not the optimal multimodal learning setting. On the contrary, we prove that imbalanced optimization obeying the inverse of modality variance ratio can contribute to optimal performance. This provides an insightful view to study under-optimized multimodal learning.
    % \item We introduce the modality variance metric and prove that modality dependency obeying the modality variance contributes the optimal performance. This provides an insightful view to study under-optimized multimodal learning.
    \item We propose the ARL strategy to modulate the gradients of each modality, accelerating their optimization and aligning their contribution to model optimization with their variance.
    \item Extensive experiments on various tasks and datasets demonstrate the effectiveness and generalization of the proposed ARL in boosting multimodal learning.
\end{itemize}

%-------------------------------------------------------------------------

\section{Related Works}
\subsection{Multimodal Learning}
Models fusing data from multiple modalities have shown superior performance over unimodal models in various applications, such as action recognition~\cite{ar1,ar2}, object detection~\cite{mm_detection2,mm_detection3,wei2023mshnet}, and audiovisual speech recognition~\cite{av1,av2}. Researchers have pursued various research directions based on specific applications. For instance, some have delved into investigating the fusion strategy and network framework to improve the model's robustness to incomplete multimodal data~\cite{imc2, wei2023mmanet, wei2025robust}.  Furthermore, substantial efforts have been dedicated to harnessing information from multiple modalities to boost model performance in specialized tasks. These tasks encompass action recognition~\cite{MH3,MARS,MH2}, semantic segmentation~\cite{rgbd_seg1,rgbd_seg2,rgbd_seg3} and audio-visual speech recognition~\cite{av1,av2}. Nonetheless, most multimodal methods employing joint training strategies could not fully exploit all modalities and produce under-optimized unimodal representations. Consequently, the performance of multimodal models falls short of expectations, and even cannot achieve better performance compared to the best single-modal DNNs~\cite{wh}.

\subsection{Under-optimized Multimodal Learning}

The aforementioned defect of multimodal learning methods encourages researchers to explore the reasons behind it. Wang~\etal~\cite{wh} found that different modalities overfit and generalize at different rates and thus obtain suboptimal solutions. To this end, they calculate the logit output of each modality and their fusion and then optimize the gradient mixing problem to obtain better weights for each branch.

Furthermore, some works~\cite{umt,ogm,pmr,mmcosine,agm,MLA,FMV} proposed that the better-performing modality will dominate the gradient update while suppressing the learning process of the other modality. To this end, Du~\etal~\cite{umt} aimed to enhance unimodal performance by distilling knowledge from well-trained models. However, this introduces more model structure and computational effort, making the training process more complex and expensive. Therefore, OGM\cite{ogm} chose to reduce the gradient of the dominant modality to mitigate its inhibitory effect on the other modalities. Xu~\etal~\cite{mmcosine} then extended OGM to the multi-modal fine-grained task via the modality-wise L2 normalization to features and weights. Although a certain degree of improvement is achieved, such approaches do not impose the intrinsic motivation of improvement on the slow-learning modality. Thus, Fan~\etal~\cite{pmr} introduced the prototypical rebalance strategy to accelerate the slow-learning modality and alleviate the suppression of the dominant modality. Furthermore, MLA~\cite{MLA} and ReconBoost~\cite{reconboost} transformed the vanilla joint multimodal learning process into an alternating unimodal process to minimize inter-modality interference directly. MMPareto~\cite{mmpareto} introduced Pareto integration to alleviate the gradient conflict between multimodal and unimodal learning objectives. D\&R~\cite{diagnosing} estimated the separability of each modality in its unimodal representation space, and then used it to softly re-initialize the corresponding uni-modal encoder

\begin{figure*}[ht]
\centering
\includegraphics[width=1.0\textwidth]{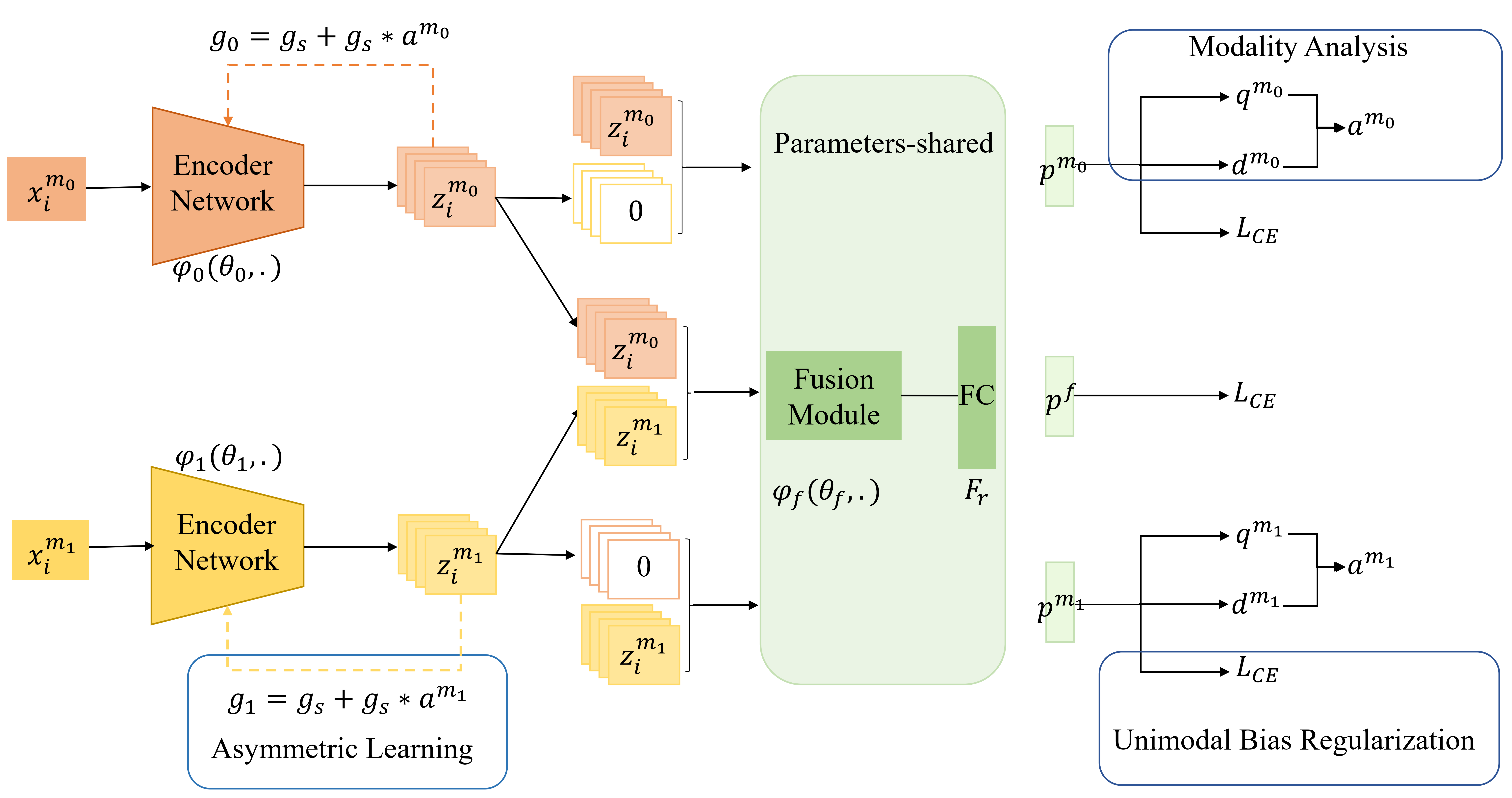}
\caption{ The pipeline of the multimodal model with ARL strategy. It consists of three components: the modality analysis to measure the modality variance; the asymmetric learning to align the optimization dependency on each modality with the inverse of their variance ratio; and the unimodal bias regularization to reduce multimodal bias. $g_s$ is the gradient passed back from the fusion module.}
\label{framework}
% \vspace{-0.2cm}
\end{figure*}

% $g_o$ and $g_p$ denote the gradient of optimization-dominant and performance-dominant modalities, respectively. $g_{r}$ denotes the gradient for the remaining modality

Generally, these methods aim to improve multimodal learning via unimodal assistance or balanced learning. This considers all modalities to be equally important, ignoring their inherent capacity discrepancy. This paper argues that balanced optimization is not the optimal setting for multimodal learning.  On the contrary, imbalanced learning obeying modality variance can contribute to better performance.

% Generally, these methods attribute the performance bottleneck of multimodal learning to imbalanced learning, where the dominant modality suppresses the learning process of the other modality.  This paper argues that balanced optimization is not the optimal setting for multimodal learning.  On the contrary, imbalanced learning obeying modality variance can contribute to better performance.

% Different from them, this paper reveals that the under-optimized of the performance-dominant modality is exactly the real performance bottleneck, even if it is the optimization-dominant modality. More importantly, our experiments show that balanced learning of different modalities is not the optimal multimodal learning setting. On the contrary, letting the performance-dominant modality dominate the learning can contribute a better performance.

% While these methods achieve promising performance, they focus on the weaker optimization modality in the training stage while ignoring the performance-dominant modality. However, our experiments show that the under-optimized of the performance-dominant modality is exactly the real reason for the multimodal performance bottleneck. Besides, they 

\section{Method}

% Existing balance-based methods attribute insufficient multimodal performance to imbalanced learning across modalities and seek to rectify this through modality rebalancing. However, this paper contends that this may be sub-optimal. as shown in Fig.~\ref{diveristy-compare} (b), when we increase the gradient of the perform-better modality by 5 times, enlarging the training imbalance, the model performance does not decrease but increases.  This motivates us to explore the real bottleneck that limits multimodal learning performance. In the following.

We will first re-analyze the under-optimized problem in multimodal learning and prove that a multimodal model can achieve optimal performance when the optimization dependencies on each modality are consistent with the inverse of their variance ratio. Then we introduce the ARL strategy to boost multimodal learning according to this theory.

% a novel insight to inferior multimodal performance via  re-analyze the Under-optimized Problem in Multimodal Learning

\subsection{Re-analyze the Under-optimized Problem in Multimodal Learning.}

We will first introduce the basic model for multimodal learning. Then we re-analyze the under-optimized phenomenon in multimodal learning.

% We introduce the analysis of the under-optimized problem for multimodal learning and find that the performance bottleneck of multimodal learning is caused by the under-optimized of performance-dominant modalities rather than the weak optimization modality. 

% \subsubsection{formulation}

\textbf{Multimodal learning model.} Without loss of generality, we consider two input modalities as $m_0$ and $m_1$. The dataset is denoted as $\mathcal{D}=\left\{x_i^{m_0}, x_i^{m_1}, y_i\right\}_{i=1,2, \ldots, N}$, where $y \in \{1,2, \ldots, M\}$, and $M$ is the number of categories. We use two encoders $\varphi_{0}\left(\boldsymbol{\theta}_{0}, \cdot\right)$ and $\varphi_{1}\left(\boldsymbol{\theta}_{1}, \cdot\right)$ to extract features, where $\boldsymbol{\theta}_0$ and $\boldsymbol{\theta_1}$ are the parameters of encoders. 
The representation outputs of encoders are denoted as $\boldsymbol{z}_0=\phi_0\left(\boldsymbol{\theta}_{0}, {{x}_{i}}^{m_0}\right)$ and $\boldsymbol{z}_1=\phi_1\left(\boldsymbol{\theta}_{1}, {{x}_{i}}^{m_1}\right)$. The two unimodal encoders are connected through the representations by some kind of fusion methods, which is prevalent in multimodal learning~\cite{MARS,tensor-1,IMC}. Here, let $\phi_{f}(\boldsymbol{\theta_{f}},.)$ denotes the fusion module. $\boldsymbol{\theta}_{f}$ is the parameter of this module. Let $\boldsymbol{W} \in \mathbb{R}^{M \times\left(d_{0}+d_{1}\right)}$ and $ \boldsymbol{b} \in \mathbb{R}^M$ denote the parameters of the linear classifier to produce the logits output. The output of input $x_{i}$ in a multimodal model  can be expressed as follows,
\begin{numcases}{}
    f({x}_{i})=\boldsymbol{W} \boldsymbol{z}_{f}+\boldsymbol{b} \\
    \label{zf}
    \boldsymbol{z}_{f}= \phi_{f}(\boldsymbol{\theta_{f}},\boldsymbol{z}_0;\boldsymbol{z}_1)
\end{numcases}

% Here $\boldsymbol{W}$ and $\boldsymbol{b}$ can be divided into the combination of  $\boldsymbol{W}_0 \in \mathbb{R}^{M \times d_{z_0}}, \boldsymbol{W}_1 \in \mathbb{R}^{M \times d_{z_1}}$ and $\boldsymbol{b}_0, \boldsymbol{b}_1 \in \mathbb{R}^M$, respectively. And $ f({{x}_{i}})$ can be rewrite as follows,

% \begin{equation}
% \begin{aligned}
% f({{x}_{i}}) & =\boldsymbol{W}_0 \cdot \boldsymbol{z}_0+\boldsymbol{b}_0  +\boldsymbol{W}_1 \cdot \boldsymbol{z}_1+\boldsymbol{b}_1
% \end{aligned}
% \end{equation}

Here, take the most widely used vanilla fusion method, concatenation, as an example, $\boldsymbol{z}_f=[\boldsymbol{z}_0;\boldsymbol{z}_1]$ and thus $f(x_i)$ can be rewritten as follows,

\begin{equation}
\begin{aligned}
\label{decouple}
f({{x}_{i}}) & =\boldsymbol{W}_0 \cdot \boldsymbol{z}_0+\boldsymbol{b}_0  +\boldsymbol{W}_1 \cdot \boldsymbol{z}_1+\boldsymbol{b}_1
\end{aligned}
\end{equation} where $\boldsymbol{W}=[\boldsymbol{W}_{0};\boldsymbol{W}_1]$, $\boldsymbol{W}_0 \in \mathbb{R}^{M \times d_{0}}, \boldsymbol{W}_1 \in \mathbb{R}^{M \times d_{1}}$ and $\boldsymbol{b}_0, \boldsymbol{b}_1 \in \mathbb{R}^M$, respectively.

Let $\boldsymbol{s}^{m_0}_{i}=\boldsymbol{W}_0 \cdot \boldsymbol{z}_0+\boldsymbol{b}_0 $, which denotes the logit output of modality $m_0$, and $\boldsymbol{s}^{m_1}_{i}=\boldsymbol{W}_1 \cdot \boldsymbol{z}_1+\boldsymbol{b}_1 $, which denotes the logit output of modality $m_1$. As a result, the final output is the summation of $\boldsymbol{s}^{m_0}_{i}$ and $\boldsymbol{s}^{m_1}_{i}$. Thus, the gradient is determined by $\boldsymbol{s}^{m_0}_{i}$ and $\boldsymbol{s}^{m_1}_{i}$ and the optimization dependency coefficient $d$ is defined as follows,
\begin{equation}
\label{impact}
 d=\frac{d^{m_0}}{d^{m_1}}=\frac{exp(\boldsymbol{s}^{m_0}_{i}(y_i))/\sum_{j=1}^{M} exp(\boldsymbol{s}^{m_0}_{i}(j))}{exp(\boldsymbol{s}^{m_1}_{i}(y_i))/\sum_{j=1}^{M} exp(\boldsymbol{s}^{m_1}_{i}(j))}
\end{equation}

% where $|.|$ means the absolution operator.

Existing studies~\cite{ogm,pmr,agm} hold that the modality with better performance will dominate the optimization of the weak one and lead to insufficient multimodal learning. Thus they try to rebalance the learning of different modalities by gradient modulation to make $d$ equal 1.

However, as shown in Fig.~\ref{diveristy-compare} (b), when we increase the gradient of the perform-better modality by 5 times, enlarging the training imbalance, the model performance does not decrease but increases.  This finding drives us to further explore the real bottleneck that limits multimodal learning performance. More specifically, \textit{what is the optimal optimization dependency ratio for different modalities?}

\textbf{The optimal optimization dependency ratio for different modalities}. As discussed above, balancing the contribution of different modalities to multimodal model optimization is not optimal. This drives us to find the optimal contribution ratio of different modalities. Generally, the optimal contribution ratio should minimize the generalization error, thus we formulate the problem as follows,

\begin{numcases}{}
\label{target}
    \mathop{\min}_{w_0,w_1} g=\frac{1}{N}\sum_{i=1}^{i=N}L(f(x_i),y_i) \\
    f(x_i)=w_0 {s}^{m_0}_{i} +w_1 {s}^{m_1}_{i} 
\end{numcases} where $w_0>0$ and $w_1>0$ denotes the contribution of modality $m_0$ and $m_1$, respectively, and $w_0+w_1=1$. $L(.)$ measures the generalization error from prediction and groundtruth. For simplification, we define $E\left[f(x)\right]=\frac{1}{N}\sum_{i=1}^{i=N}f(x_i)$.According to the bias-variance decomposition~\cite{bias}, the error between prediction and groundtruth can be rewritten as follows,

\begin{numcases}{}
g= (Bias(f(x),y))^{2}+Var(f(x))+Var(\epsilon) \\
 Bias(f(x),y)=E\left[f(x)-y\right] \\
 Var(f(x))=E\left[f(x)^2\right]-E\left[f(x)\right]^2
\end{numcases} where $Bias(f(x),y)$ is the bias of prediction, which measures the error of the prediction. $Var(f(x))$ is the variance of $f(x)$, which measures the uncertainty of the prediction, $Var(\epsilon)$ is the irreducible error in the dataset and cannot be reduced by any model. In other words,  to minimize the Eq.~\ref{target}, we need to minimize the $Bias(f(x),y)^2$ and $Var(f(x))$. Limited by page, we only give the key results here and detailed derivation can be seen in the supplementary materials.

% \begin{equation}
%     \frac{1}{N}\sum_{i=1}^{i=N}(L(f(x_i),y_i))^{2}=\frac{1}{N}\sum_{i=1}^{i=N} (Bias(f(x_i),y_i))^{2}+\frac{1}{N}\sum_{i=1}^{i=N}Var(f(x_i))+Var(\epsilon)
% \end{equation} where $Bias(f(x_i),y_i)$ is the bias of prediction, which measures the accuracy of the prediction. $Var(f(x_i))$ is the variance of $f(x_i)$, which measures the uncertainty of the prediction, $Var(\epsilon)$ is the irreducible error in the dataset and cannot be reduced by any model. 

For the term of $Bias(.)^2$, we can convert it as follows,

\begin{equation}
\begin{aligned}
Bias(f(x),y)^2
&=({w_0}Bias({s}^{m_0},y)+{w_1}Bias(({s}^{m_1},y))^2
\end{aligned}
\end{equation}
% &=E\left[w_0 ({s}^{m_0}-y) +w_1 ({s}^{m_1}-y)\right]\\
 % since $w_0+w_1=1$,

% \begin{numcases}{}
% Bias(f(x),y)^2&=&E\left[w_0 ({s}^{m_0}_{i}-y) +w_1 ({s}^{m_1}_{i}-y)\right]\\
% &=&({w_0}Bias(f(x^{m_0}),y)+{w_1}Bias(f(x^{m_1}),y))^2
% \end{numcases} where $f(x^{m_0}={s}^{m_0}$ and $f(x^{m_1}={s}^{m_1}$. The derivation details can be seen in the appendix.

Then, with the constraint $w_0+w_1=1$, we get the solution of $w_0$ and $w_1$ as follows,

 % according to the Lagrange multiplier method,

\begin{numcases}{}
    w_0=\frac{Bias({s}^{m_1},y)}{Bias({s}^{m_1},y)-Bias({s}^{m_0},y)}\\
    w_1=\frac{-Bias({s}^{m_0},y)}{Bias({s}^{m_1},y)-Bias({s}^{m_0},y)}
\end{numcases} Here, the numerical solution is meaningless since one of $w_0$ or $w_1$ must be smaller than $0$, conflicting with $w_1>0$ and $w_1>0$. Thus, when $Bias({s}^{m_0},y)$ and $Bias({s}^{m_1},y)$ are fixed, we can not find a reasonable combination of $w_0$ or $w_1$ to minimize $Bias(.)^2$. Consequently, the only way to minimize $Bias(.)^2$ is to minimize $Bias({s}^{m_0},y)$ and $Bias({s}^{m_1},y)$.

% Thus  However, we can get the relationship between optimal $w_0$ as well as $w_1$ and $Bias({s}^{m_0},y)$ as well as $Bias({s}^{m_1},y)$ as follows,

% \begin{equation}
% \label{prop}
%      \frac{w_0}{w_1}\propto \frac{\frac{1}{Bias({s}^{m_0},y)}}{\frac{1}{Bias({s}^{m_1},y)}}  
% \end{equation}. In other words, to minimize the $Bias(f(x),y)^2$, we should enable the modality with a lower prediction bias to have a larger contribution to model optimization.

For the term of $Var(f(x))$, we can convert it as follows,
% &=Var(w_0{s}^{m_0}+w_1{s}^{m_1})\\
\begin{equation}
\begin{aligned}
    Var(f(x))
    &={w_0}^2Var({s}^{m_0})+{w_1}^2 Var({s}^{m_1})
\end{aligned}
\end{equation} Then, with the constraint $w_0+w_1=1$, we get the solution of $w_0$ and $w_1$ as follows,
% , according to the Lagrange multiplier method,
\begin{numcases}{}
    w_0=\frac{Var({s}^{m_1})}{Var({s}^{m_1})+Var({s}^{m_1})}\\
    w_1=\frac{Var({s}^{m_0})}{Var({s}^{m_1})+Var({s}^{m_1})}
\end{numcases}. Here,we can get the relationship between $\frac{w_0}{w_1}$ and $Var({s}^{m_0})$ as well as $Var({s}^{m_1})$ as follows,

\begin{equation}
     \frac{w_0}{w_1} = \frac{\frac{1}{Var({s}^{m_0})}}{\frac{1}{Var({s}^{m_1})}}  
\end{equation}. In other words, to minimize the $Var(f(x))$ of multimodal models,  we should enable the modality with a lower variance to have a larger contribution to model optimization.
%
% the modality with a lower uncertainty should have a larger contribution to model optimization.

% Therefore, to minimize Eq.~\ref{target}, the optimal contribution dependency ratio for $m_0$ and $m_1$ should be,

% \begin{equation}
% \label{opt}
%   \frac{w_0}{w_1} = k \frac{\frac{1}{Bias({s}^{m_0},y)Var({s}^{m_0})}}{\frac{1}{Bias({s}^{m_1},y)Var({s}^{m_1})}} 
% \end{equation} where $k$ is a number larger than $0$, representing the proportional relationship in Eq.~\ref{prop}. For simplification, we set k $k=1$. Here, we define the $\frac{1}{Bias({s}^{m_0},y)Var({s}^{m_0})}$ as the variance factor $q^{m_0}$ of modality $m_0$. Consequently, the modality with low prediction bias and uncertainty is a high-variance modality, which is consistent with the intuition.

\subsection{Asymmetric Representation Learning}
\label{arl}

As discussed, balanced learning is not optimal. In contrast, imbalanced learning inversely proportional the modality variance ratio can contribute to better performance. Moreover, because there is no reasonable combination to minimize the fusion bias, we need to minimize the unimodal bias to achieve a lower generalization error. To this end, we propose a simple but effective Asymmetric Representation Learning (ARL) strategy to improve the performance of multimodal learning. As shown in Fig~\ref{framework}, it consists of three components: modality analysis, asymmetric learning, and unimodal bias regularization. 

% Besides, the pseudocode of multimodal learning with ARL is given in  Algorithm~\ref{dgd}.

% we get two criteria. 1)All modality in multimodal learning is under-optimized compared to unimodal learning, including the one dominating the optimization. 2)More importantly, balancing the learning of different modalities is not the optimal setting for multimodal learning. On the contrary, we should guide the performance-dominant modality to dominate the optimization of multimodal learning.

 % as shown in Ar~\xxxx

\textbf{Modality Analysis} This stage determines the inverse of the variance $q$ of each modality. According to the definition, we need to calculate their logit output first. To this end, we introduce auxiliary regularizers for each modality to calculate their logit output directly. Compared to using $s^{m_0}$ and $s^{m_1}$ calculated in vanilla concatenation fusion, this is independent of model structures and fusion methods, making it highly versatile for complex scenarios. Notably, As shown in Fig~\ref{framework}, this introduces no extra parameters since all regularizers share the parameters with the multimodal model and calculate the unimodal loss by setting other modality representations to zero.

 % ${s}^{m_0} $ and ${s}^{m_1}$ from

% make some modifications to

Since the vanilla bias-variance decomposition is designed for the regression task, we need to make it suitable for the classification task. Following existing work~\cite{gupta2022ensembles,zhang2024multimodal}, we adapt the self-information entropy to approximate the variance term in the classification task. Take the modality $m_0$ as an example, we measure $q^{m_0}$ as follows,

\begin{equation}
\label{qq}
  q^{m_0}=1/\sum_{j=1}^{j=M}\sigma(p^{m_0})(j)log\sigma(p^{m_0})(j)
\end{equation}{} where $p^{m_0}$ and $p^{m_1}$ is the logit output of modality $m_0$ and $m_1$, respectively. $\sigma(.)$ is the softmax function.

\textbf{Asymmetric Learning} This stage calculates the modulation coefficient $a^{m_0}$ and $a^{m_1}$ for each modality encoder to adjust their modality contribution ratio so that it can be inversely proportional to the modality variance ratio. 

% Then we regulate the optimization processing via a residual link.

To make the optimization dependency ratio $\frac{d^{m_0}}{d^{m_1}}$ equal the modality variance ratio $\frac{q^{m_0}}{q^{m_1}}$, we should increase the gradient of $m_0$ to increase $\frac{d^{m_0}}{d^{m_1}}$ when it is smaller than $\frac{q^{m_0}}{q^{m_1}}$. Or we should increase the gradient of $m_1$ to decrease $\frac{d^{m_0}}{d^{m_1}}$  when it is larger than $\frac{q^{m_0}}{q^{m_1}}$. Therefore, we define the modulation coefficient $a^{m_0}$ and $a^{m_1}$ as follows,

% the modality dependency ratio when the modality variance ratio is smaller than the current modality dependency ratio. Therefore, we define the modulation coefficient $a^{m_0}$ and $a^{m_1}$ as follows,

\begin{equation}  
\left[a^{m_0},a^{m_1}\right]=\sigma(\left[\frac{q^{m_0}}{q^{m_1}}*T,\frac{d^{m_0}}{d^{m_1}}*T\right])
\end{equation} where $\frac{d^{m_0}}{d^{m_1}}$ are calculated by Eq.~\ref{impact} with $s^{m_0}=p^{m_0}$ and $s^{m_1}=p^{m_1}$. This represents the real modality dependency ratio in the current stage. $\sigma(.)$ is the softmax function. $T$ is the temperature coefficient, which controls the intensity of the modulation. Larger $T$ will increase the gap between $a^{m_0}$ and $a^{m_1}$, accelerating the modulation. Thus the gradient $g_s$ from shared modules passed back to modality encoder $m_0$ and $m_1$ are regulated as follows,

\begin{numcases}{}
{g}^{m_0}_{s}=g_s+g_s*a^{m_0}\\
{g}^{m_1}_{s}=g_s+g_s*a^{m_1}
\end{numcases} where the residual sum with the original gradient is employed to prevent the parameter updating from stopping when the modulation factor is extremely small.

\textbf{Unimodal Bias Regularization}  Since there is no reasonable combination that minimizes the fusion bias, this stage ARL integrates the prediction bias of each modality and optimizes them in conjunction with the multimodal loss to further reduce the generalization error.

Take the modality $m_0$ as an example, we measure its bias $u^{m_0}$ from its logit output as follows,

\begin{equation}
  u^{m_0}=L_{CE}(p^{m_0},y)
\end{equation} where $L_{CE}$ is the cross entropy loss.

\textbf{Total loss}. As discussed above, the total loss $L_{ARL}$ for multimodal learning with ARL is defined as follows,

\begin{equation}
   L_{ARL}=L_{CE}(p^{f},y)+\gamma(u^{m_0}+u^{m_1})
\end{equation} where $p^{f}$ is the multimodal logit output. $L_{CE}(p^{f},y)$ optimize the prediction error of the multimodal model. $u^{m_0}+u^{m_1}$ optimize the unimodal bias of the multimodal model. The training pseudocode of the multimodal model with ARL is given below,

\begin{algorithm}[h]
	\caption{Multimodal learning with ARL strategy} 
	\label{dgd} 
	\begin{algorithmic}
		\REQUIRE Training dataset D, iteration number T.
        \STATE \textbf{for} t = 0, · · · , T \textbf{do}
        % \STATE  \textbf{Unimodal Bias Regularization and Modality Analysis}
        \STATE \quad\quad Feed-forward the batched data to the model.
        \STATE \quad\quad Calculate unimodal and multimodal loss via Eq.22.
        \STATE \quad\quad Calculate backward gradient.
        \STATE \quad\quad Calculate the variance of each modality via Eq.17.
        \STATE \quad\quad Calculate the modulation coefficient via Eq.18.
        \STATE \quad\quad Modify the gradient of each modality via Eq.19.
        \STATE \quad\quad Update model parameters via the modified gradient.
	\end{algorithmic} 
\end{algorithm}

\subsection{Comparison with Existing Method}

Existing balance-based methods~\cite{ogm,agm,pmr,mmpareto} try to modulate the learning of different modalities to balance their optimization. However, we found that imbalanced learning obeying the inverse of the modality variance ratio can contribute to better performance. Therefore, we proposed ARL technology to modulate the learning of different modalities, forcing their contribution to model optimization to satisfy the inverse of their variance ratio.

% when increasing the learning of weak modalities.

% learning strength and modal variance consistent between different modalities. That is, giving higher weight to high-variance modalities.
% This ignores their specificity and leads to insufficient representation learning.

\section{Experiments}
\subsection{Datasets}
\textbf{CREMA-D}~\cite{Crema} is an audio-visual dataset for emotion recognition, which consists of audio and visual modalities. There are a total of 7442 videos in the dataset for 6 usual emotions. They are further divided into 6698 samples as the training set and 744 samples as the testing set.

\textbf{AVE}~\cite{AVE} is an audio-visual video dataset for audio-
visual event localization. It comprises 4,143 10-second videos across 28 event classes. Here, we extract frames from event-localized video segments and capture corresponding audio clips, creating a labeled multimodal classification dataset. The training and validation split of the dataset follows~\cite{AVE}.

\textbf{Kinetics-Sounds (KS)}~\cite{ks} is a dataset derived from the Kinetics dataset, focusing on 34 human action classes. Each class is chosen to be potentially manifested visually and aurally. This dataset consists of 19k 10-second video clips, with a distribution of 15k for training, 1.9k for validation, and 1.9k for testing.

\textbf{MOSI}~\cite{mosi}  is a well-known dataset in the field of multimodal sentiment analysis, encompassing audio, visual, and textual data. It comprises 2,199 video clips of individual utterances, extracted from 93 monologue videos. Each clip is annotated with a sentiment score on a continuous scale from -3 (strongly negative sentiment) to 3 (strongly positive sentiment). In this paper, we use the two-class label setting that considers positive/negative results only. We employ this dataset to demonstrate the method’s capability to generalize in environments involving multiple modalities.

\textbf{UCF-101}~\cite{soomro2012ucf101} is a multimodal dataset for human action recognition, offering both RGB and optical flow sequences. This dataset comprises 101 distinct action categories.  According to its original protocol, 9,537 video samples are used for training and 3,783 for testing. 
% \textbf{VGGSound}~\cite{vggsound} is a video dataset with 309 categories, capturing various audio events in daily life. For our experiment, we employed 168,618 videos for training and validation, and 13,954 videos for testing. This helps study the effectiveness of ARL to the large-scale dataset.

% \textbf{CEFA}~\cite{cefa} is a multimodal dataset build for face anti-spoofing task. Here, we use the RGB and Depth modality for experiments to demonstrate the effectiveness of ARL beyond the audio-visual tasks. We follow the cross-ethnicity and cross-attack protocol suggested by the authors and divide it into train, validation, and test sets with 35k, 18k, and 54k samples, respectively.

% We employ this dataset to demonstrate the effectiveness of our method beyond the audio-visual tasks.

% is a synthetic dataset derived from the MNIST dataset [18]. It contains two distinct images for each instance: a grayscale image and a monochromatic image. There are  60k instances in the training set, of which the monochromatic images exhibit a strong relation with their labels. Besides, there are 10k instances in the validation set, of which the monochromatic images possess a weaker correlation with their labels. In this study, we consider the monochromatic image as the first modality and the grayscale image as the second modality, aligning with the setup employed in a previous study~\cite{pmr}.

\subsection{Experimental settings}

\textbf{Implementation details.} To ensure a fair comparison, we utilize ResNet18 as the encoder backbone for all datasets, as is common in previous methods~\cite{ogm,pmr}. For CREMA-D, we select one frame from each clip and resize it to 224x244 as the visual input. The audio data is converted into a spectrogram of size 257×299 using librosa~\cite{librosa}. For AVE and Kinetics-Sounds datasets, we uniformly sample 3 frames from each video clip and resize them to 224x224 as visual inputs. The entire audio data is transformed into a spectrogram of size 257×1,004. The training configurations mirror those of previous methods~\cite{ogm,pmr}, including a mini-batch size of 64, an SGD optimizer with momentum 0.9, a learning rate of 1e-3, and a weight decay of 1e-4. $\gamma$ is set as 4 for all datasets. $T$ is set 8,4,4,4,4 for CREMA-D, AVE, KS, MOSI, and UCF101 datasets, respectively. More ablation studies are provided in the supplementary material.

% $T$ is set as 8 for all experiments. 

% All the analyses are conducted on PyTorch1.13 with two NVIDIA GeForce GTX3090.

% for  CREMA-D, AVE, and KS, CEFA

\textbf{Comparison settings.} To study the advantage of ARL, we make comparisons with three gradient modulation approaches, OGM~\cite{ogm}, AGM~\cite{agm}, PMR~\cite{pmr}, and four unimodal regularization methods, G-Blending~\cite{wh}, MLA~\cite{MLA}, MMPareto~\cite{mmpareto} and D\&R~\cite{diagnosing}. For a fair comparison, we unify the backbone as ResNet18 and the fusion method as concatenation in all experiments.  Note that the original MLA uses epoch as 150 and batch size as 16. We unify them with other comparison methods as epoch 100 and batch size 64. Besides, the learning rate of vanilla MMPareto~\cite{mmpareto} and D\&R~\cite{diagnosing} is 2e-3, we also unify it as  1e-3.

\subsection{Comparison on multimodal tasks}

\begin{table*}[]
\centering
\renewcommand\arraystretch{1.1}
\setlength{\tabcolsep}{4mm}{
\begin{tabular}{c|cc|cc|cc}
\hline
Dataset       & \multicolumn{2}{c|}{CREMA-D}                         & \multicolumn{2}{c|}{KS}                              & \multicolumn{2}{c}{AVE}                             \\ \hline
Methods       & \multicolumn{1}{c}{Acc}            & macro F1       & \multicolumn{1}{c}{Acc}            & macro F1       & \multicolumn{1}{c}{Acc}            & macro F1       \\ \hline
Audio-only    & \multicolumn{1}{c}{57.27}          & 57.89          & \multicolumn{1}{c}{48.67}          & 48.89          & \multicolumn{1}{c}{62.16}          & 58.54          \\ 
Visual-only   & \multicolumn{1}{c}{62.17}          & 62.78          & \multicolumn{1}{c}{52.36}          & 52.67          & \multicolumn{1}{c}{31.40}          & 29.87          \\ \hline
Concatenation & \multicolumn{1}{c}{58.83}          & 59.43          & \multicolumn{1}{c}{64.97}          & 65.21          & \multicolumn{1}{c}{66.15}          & 62.46          \\
Grad-Blending & \multicolumn{1}{c}{68.81}          & 69.34          & \multicolumn{1}{c}{67.31}          & 67.68          & \multicolumn{1}{c}{67.40}          & 63.87          \\ 
OGM-GE        & \multicolumn{1}{c}{64.34}          & 64.93          & \multicolumn{1}{c}{66.35}          & 66.76          & \multicolumn{1}{c}{65.62}          & 62.97          \\ 
AGM           & \multicolumn{1}{c}{67.21}          & 68.04          & \multicolumn{1}{c}{65.61}          & 65.99          & \multicolumn{1}{c}{64.50}          & 61.49          \\ 
PMR           & \multicolumn{1}{c}{65.12}          & 65.91          & \multicolumn{1}{c}{65.01}          & 65.13          & \multicolumn{1}{c}{63.62}          & 60.36          \\ 
MMPareto      & \multicolumn{1}{c}{70.19}          & 70.82          & \multicolumn{1}{c}{69.13}          & 69.05          & \multicolumn{1}{c}{68.22}          & 64.54          \\ 
MLA           & \multicolumn{1}{c}{73.21}          & 73.77          & \multicolumn{1}{c}{\underline{69.62}}          & \underline{69.98}          & \multicolumn{1}{c}{\underline{70.92}}          & \underline{67.23}          \\
D\&R          & \multicolumn{1}{c}{\underline{73.52}}          & \underline{73.96}          & \multicolumn{1}{c}{69.10}          & 69.36          & \multicolumn{1}{c}{69.62}          & 64.93          \\ \hline
ARL           & \multicolumn{1}{c}{\textbf{76.61}} & \textbf{77.14} & \multicolumn{1}{c}{\textbf{74.28}} & \textbf{74.03} & \multicolumn{1}{c}{\textbf{72.89}} & \textbf{68.04} \\ \hline
\end{tabular}}
\caption{Comparison with existing modulation strategies on CREMA-D, Kinetics-Sounds, and AVE datasets. The proposed ARL achieves the best performance. Bold and underline mean the best and second-best results, respectively.}
\label{compare}
\end{table*}

\begin{table}[]
\centering
\begin{tabular}{c|cc|cc}
\hline
Dataset       & \multicolumn{2}{c|}{MOSI}                            & \multicolumn{2}{c}{UCF101}                          \\ \hline
Methods       & \multicolumn{1}{c}{Acc}            & macro F1       & \multicolumn{1}{c}{Acc}            & macro F1       \\ \hline
Concatenation & \multicolumn{1}{c}{76.92}          & 75.68          & \multicolumn{1}{c}{80.41}          & 79.40          \\ 
Grad-Blending & \multicolumn{1}{c}{78.16}          & 76.94          & \multicolumn{1}{c}{81.73}          & 80.84          \\ 
OGM-GE        & \multicolumn{1}{c}{77.33}          & 76.23          & \multicolumn{1}{c}{81.15}          & 80.36          \\ 
AGM           & \multicolumn{1}{c}{77.65}          & 76.49          & \multicolumn{1}{c}{81.55}          & 80.36          \\ 
PMR           & \multicolumn{1}{c}{77.48}          & 76.34          & \multicolumn{1}{c}{81.36}          & 80.37          \\
MMPareto      & \multicolumn{1}{c}{78.17}          & 76.92          & \multicolumn{1}{c}{81.98}          & 80.64          \\ 
MLA           & \multicolumn{1}{c}{78.87}          & 77.21          & \multicolumn{1}{c}{82.01}          & \underline{81.22}          \\ 
D\&R          & \multicolumn{1}{c}{\underline{78.96}}          &\underline{ 77.37 }         & \multicolumn{1}{c}{\underline{82.11}}          & 80.87          \\ \hline
ARL           & \multicolumn{1}{c}{\textbf{79.94}} & \textbf{78.84} & \multicolumn{1}{c}{\textbf{83.22}} & \textbf{81.98} \\ \hline
\end{tabular}
\caption{\textbf{(Left)}: Comparison with imbalanced multimodal learning methods on MOSI dataset with three modalities.\textbf{(Right)}: Comparison with imbalanced multimodal learning methods on the UCF101 dataset with 101 categories. Bold and underline mean the best and second-best results, respectively.}
\label{more}
\vspace{-0.7em}
\end{table}

% \begin{table}[]
% \centering
% \renewcommand\arraystretch{1.3}
% \begin{tabular}{c|c|c|c|c}
% \hline

% Dataset       & CREMA-D & AVE  & KS   & MOSI \\ \hline

% Method        & Acc     & Acc  & Acc   & Acc  \\ \hline

% Audio-only  &   57.27     &        62.16      &      48.67     &        76.4\\
% Visual-only   &   62.17      &       31.40        &    52.36        &     54.43   \\ 
% Text-only &   -      &       -       &    -       &    76.53   \\  \hline

% Concatenation & 58.83    & 66.15 & 64.97  &  76.92    \\ 
% Grad-Blending & 66.2    & 67.40 & 66.2  & 78.16 \\ 
% OGM-GE        & 64.34    & 65.62 & 66.35 &77.33 \\ 
% AGM     & 66.04    & 64.61 & 65.52& 77.65 \\ 
% PMR           & 65.17    & 63.62 & 65.25  &77.48 \\ 
% MMPareto &68.19 &67.21 &68.93  &78.17               \\ 
% MLA &\underline{73.21} &69.12 &\underline{71.92}  &78.87               \\ 
% D\&R &72.24 &\underline{69.62} &69.39  &\underline{78.96}               \\ \hline

% ARL           & \textbf{75.26}    & \textbf{71.90} & \textbf{74.39} & \textbf{79.94} \\ \hline
% \end{tabular}

% \end{table}

% To demonstrate the advantage of ARL, we make comparisons with four modulation approaches, Gradient-Blending~\cite{wh}, OGM~\cite{ogm}, MMCosine~\cite{mmcosine}, PMR~\cite{pmr} and recent SOTA method modality valuation~\cite{FMV}. For a fair comparison, we unify the backbone as ResNet18 and the fusion method as concatenation in all experiments.

\textbf{Comparison with other modulation strategies.}  The results are shown in Table~\ref{compare}. Firstly, the proposed ARL strategy achieves the best performance on all datasets with a significant improvement compared to existing methods. Compared to the second-best method, it improves the performance by 3.02\%, 2.28\%, 3.01\%, and 0.98\% on CREMA-D, AVE, KS, respectively. These results demonstrate the effectiveness of the ARL strategy.

% mainly only focus on the case of two modalities~\cite{ogm,agm,pmr}, limiting their applicability to broader, more complex scenarios. In contrast, ARL imposes no restrictions on the number of modalities, allowing for greater flexibility. Here, we conduct experiments on the MOSI dataset with three modalities: audio, vision, and text. For a comprehensive comparison, we retain the core uni-modal balancing strategy of G-Blending, OGM-GE, AGM, PMR, and MMPareto and extend them to more than two modality cases. As shown in the left section of Table~\ref{gene}, ARL achieves the best performance in multimodal and unimodal performance,  demonstrating its flexibility in such scenarios.

More importantly, while existing balance-based methods, such as OGM-GE, AGM, and PMR, can improve the performance on the CREMA-D, KS, and MOSI datasets, they decrease the performance on the AVE dataset. This is because the balance-based methods could hinder the learning of performance-dominant modality (audio modality) when they balance the optimization of audio and visual modalities.  In contrast, the proposed ARL also improves the performance on the AVE dataset by 5.75\% compared to the concatenation baseline. This confirms the superiority to adjust the optimization dependency on each modality obeying their variance ratio. 

% to dominate multimodal learning.

% More importantly, while existing balance-based methods can improve the performance on the CREMA-D, KS, and MOSI datasets, they decrease the performance on the AVE dataset. This is because the balance-based methods could hinder the learning of performance-dominant modality (audio modality) when they balance the optimization of audio and visual modalities.  In contrast, the proposed ARL also improves the performance on the AVE dataset by 5.1\% compared to the concatenation baseline.

% More importantly, while existing balance-based methods can improve the performance on the CREMA-D, KS, and CEFA datasets, they decrease the performance on the AVE dataset. This is because the balance-based methods could hinder the learning of performance-dominant modality (audio modality) when they balance the optimization of audio and visual modalities.  In contrast, the proposed ARL also improves the performance on the AVE dataset by 5.1\% compared to the concatenation baseline.

\textbf{Comparison in more-than-two modality case}. Existing methods primarily focus on handling datasets containing two modalities ~\cite{ogm,agm,pmr}, which has certain limitations when dealing with broader scenarios. In contrast, the ARL method is not constrained by the number of modalities and demonstrates strong adaptability. In this study, we conducted experimental validation on the MOSI dataset, which includes audio, visual, and textual modalities. For a comprehensive comparison, we extend the core unimodal balancing strategies of G-Blending, OGM-GE, AGM, PMR, and MMPareto to handle cases involving more than two modalities. As shown in the left part of Table~\ref{more}, ARL also achieves the best performance on the MOSI dataset, verifying its flexibility to the more-than-two modalities case.

\textbf{Comparison in large dataset}. Considering existing datasets used in the experiments are relatively simple, we further conduct experiments on the UCF101 dataset which consists of 101 categories. The results are shown in the right part of Table~\ref{more}. We can see that ARL achieves the best result, verifying its generalizability to more complex scenes.

\subsection{Ablation Study}

% \begin{table}[h]
% \centering
% \begin{tabular}{c|cccc}
% \hline
% Methods & Vanilla & +UR   & +AL   & ARL(UR+AL)     \\ \hline
% Acc     & 58.83   & 66.20 & 70.23 & \textbf{75.06} \\ \hline
% \end{tabular}
% \caption{The ablation study of unimodal regularization (UR), asymmetric learning (AL), and gradient residual (GR).}
% \end{table}

\textbf{Generalization to different architectures.} Apart from the convolutional neural network with the late-fusion method,  transformer-based backbones using cross-modal interaction modules are also extensively applied. To validate the applicability of ARL in more scenarios, we combine it with two intermediate fusion methods MMTM~\cite{mmtm} and mmFormer~\cite{mmformer} for all the datasets. Both of them fuse information within the modality encoder. Specifically, MMTM is a CNN-based architecture that integrates intermediate feature maps from different modalities using squeeze and excitation operations. mmFormer is a transformer-based architecture, that fuses the intermediate feature maps of different modalities with the cross-attention strategy. For both methods, we use only one frame for each video on each dataset to align their input on the benchmark. Besides, we unified the backbone network as ResNet18.As illustrated in Table~\ref{artecture}, the proposed ARL demonstrates significant performance gains across various architectures, even when fusion is employed during encoder processing, suggesting its applicability in complex scenarios.

\begin{table}[]
\centering
\renewcommand\arraystretch{1.3}
\setlength{\tabcolsep}{4.5mm}{
\begin{tabular}{c|c|c|c}
\hline
Dataset  & CREMA-D & AVE  & KS    \\ \hline
Method   & Acc     & Acc  & Acc    \\ \hline
MMTM & 51.86    & 51.52 & 56.70  \\ 
MMTM† & \textbf{58.68}    & \textbf{60.88} & \textbf{63.72}  \\ \hline
mmFormer & 59.69    & 60.41 & 64.72  \\  
mmFormer† & \textbf{68.33}    & \textbf{67.14} & \textbf{68.51} \\ \hline
\end{tabular}}
\caption{Performance on different datasets with CNN-based and transformer-based architectures, respectively. † indicates ARL strategy is applied, which achieves better performance.}
\label{artecture}
\end{table}

\begin{table}[]
\centering
\renewcommand\arraystretch{1.2}
\begin{tabular}{c|c|c|c}
\hline
Dataset & CREMA-D & AVE  & KS            \\ \hline
Vanilla & 77.66    & 84.52 & 82.31          \\ \hline
ARL     & \textbf{80.01}    & \textbf{91.11} & \textbf{85.88} \\ \hline
\end{tabular}
\caption{Experiments of Swin-Base backbone network on different datasets. The fusion method is Gated fusion. `Vanilla' means conventional multimodal model. `ARL' means the multimodal model using gradient decoupling learning.}
\label{backbone}
\vspace{-0.9em}
\end{table}

\textbf{Generalization to larger backbone networks.} Considering that ResNet18 is a relatively small backbone network, we switch to Swin-Base to examine the scalability of ARL to a larger backbone. We initialize the model with the pre-trained weights on ImageNet-22k. As shown in Table~\ref{backbone},  the introduction of ARL can continue to boost the performance of the vanilla model on all datasets. This demonstrates the effectiveness of ARL to large backbone networks.

% \begin{table}[]
% \centering
% \setlength{\tabcolsep}{3.5mm}{
% \begin{tabular}{ccc}
% \hline
% Modality & Vanilla & ARL           \\ \hline
% Audio    & 55.13    & \textbf{62.33} \\ 
% Visual   & 18.24    & \textbf{68.89} \\ \hline
% \end{tabular}}
% \caption{Unimodal performance in the vanilla and ARL multimodal model on the CREMA-D dataset. `Vanilla' represents the concatenation baseline.}
% \label{inter-modality}
% \end{table}

% \begin{table}[]
% \centering
% \begin{tabular}{c|c|c|c}
% \hline
% Dataset & CREMA-D & AVE  & KS            \\ \hline
% Vanilla & 77.6    & 84.2 & 82.1          \\ \hline
% ARL     & \textbf{80.1}    & \textbf{91.1} & \textbf{85.8} \\ \hline
% \end{tabular}
% \caption{Experiments of Swin-Base backbone network on different datasets. The fusion method is Gated fusion. `Vanilla' means conventional multimodal model. `ARL' means the multimodal model using gradient decoupling learning.}
% \label{backbone}
% \end{table}

% \textbf{Generalization to larger backbone networks.} Considering that ResNet18 is a relatively small backbone network, we switch to Swin-Base to examine the scalability of ARL to a larger backbone. We initialize the model with the pre-trained weights on ImageNet-22k. As shown in Table~\ref{backbone},  the introduction of ARL can continue to boost the performance of the vanilla model on all datasets. This demonstrates the effectiveness of ARL to large backbone networks.

\begin{figure}[t]
\centering
\includegraphics[width=1.0\columnwidth]{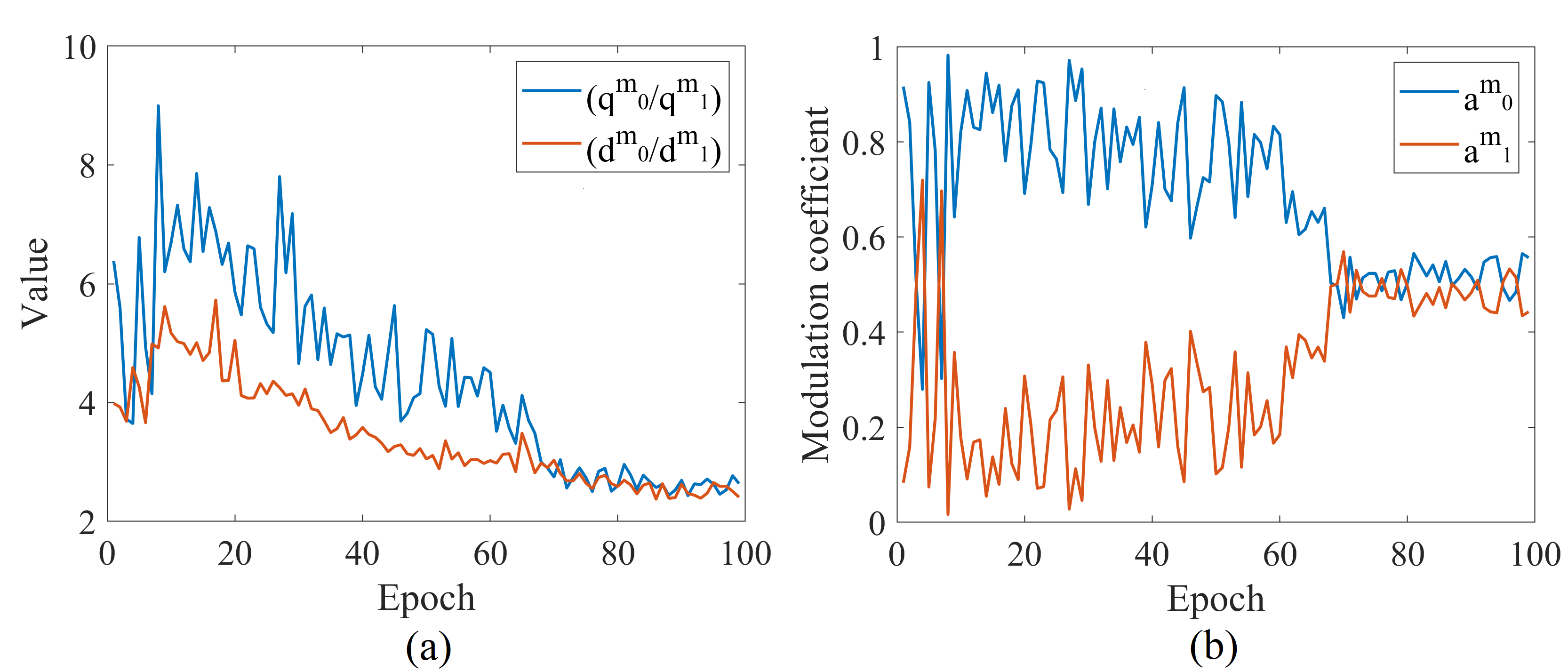}
\caption{ Visualization of the optimization dependency and modality variance ratio (a) and the corresponding modulation coefficient(b) of ARL on AVE during training. Here $m_0$ and $m_1$ represent audio and visual modalities, respectively.}
\label{weight}
% \vspace{-0.5cm}
\end{figure}

\begin{table}[]
\centering
\setlength{\tabcolsep}{4mm}{
\begin{tabular}{ccc|cc}
\hline
\multicolumn{3}{c|}{Setting}                              & \multicolumn{2}{c}{Metric}           \\ \hline
\multicolumn{1}{c}{UR}  & \multicolumn{1}{c}{AL}  & GR  & \multicolumn{1}{c}{Acc}   & macro F1 \\ \hline
\multicolumn{1}{c}{}    & \multicolumn{1}{c}{}    &     & \multicolumn{1}{c}{58.83} & 59.43    \\
\multicolumn{1}{c}{$\checkmark$} & \multicolumn{1}{c}{}    &     & \multicolumn{1}{c}{67.20} & 67.68    \\
\multicolumn{1}{c}{}    & \multicolumn{1}{c}{$\checkmark$} &     & \multicolumn{1}{c}{70.23} & 70.79    \\ 
\multicolumn{1}{c}{$\checkmark$} & \multicolumn{1}{c}{$\checkmark$} &     & \multicolumn{1}{c}{74.45} & 75.98    \\ 
\multicolumn{1}{c}{$\checkmark$} & \multicolumn{1}{c}{$\checkmark$} & $\checkmark$ & \multicolumn{1}{c}{\textbf{76.61}} & \textbf{77.14}    \\ \hline
\end{tabular}}
\caption{The ablation study of unimodal regularization (UR), asymmetric learning (AL), and gradient residual (GR) on the CREMA-D dataset. $\checkmark$ means the component is used.}
\label{comp}
% \vspace{-0.7em}
\end{table}

\textbf{The effect of each component in ARL}. We conduct experiments on the CREMA-D dataset to study the effect. The results are shown in Table~\ref{comp}. We can see that all the unimodal bias regularization (UR), asymmetric learning (AL), and gradient residual (GR) significantly improve performance over the vanilla model. UR reduces decision bias, AL reduces decision variance, and GR ensures the gradient lower bound of the suppressed modality.

\subsection{Visualization}

To understand the mechanism of ARL intuitively, we visualize the difference between optimization dependency and modality variance ratio, and the corresponding modulation coefficient during the training process of the AVE dataset. As shown in Fig.~\ref{weight} (a), the optimization dependency is lower than the modality variance ratio in the early training stage (before the 60th epoch). Therefore, as shown in Fig.~\ref{weight} (b), ARL will give $m_0$ (the audio modality) a large weight to accelerate its optimization, increasing the optimization dependency on $m_0$ to increase the $\frac{d^{m_0}}{d^{m_1}}$. Then as shown in Fig.~\ref{weight} (a), the curve of optimization dependency and modality variance will gradually converge to the same value. More importantly, we can see that the optimal ratio of optimization dependency is not 1 but 3, demonstrating the necessity of imbalanced learning.

\section{Discussion}
 \textbf{Conclusion.} In this paper, we re-analyze the under-optimized problem in multimodal learning and reveal that balanced learning is not optimal for multimodal learning. In contrast, we prove that imbalanced learning obeying the inverse of modality variance can contribute to optimal performance. Then we propose a simple but effective multimodal learning strategy called Asymmetric Representation Learning (ARL) to alleviate the under-optimized problem, accelerating their optimization and encouraging models to rely on each modality according to the inverse of their variance. This method achieves consistent performance gain on five representative multimodal datasets under various settings. In addition, ARL can also generally serve as a flexible plug-in strategy for both CNN and Transformer-based models, demonstrating its practicality.

 \textbf{Limitation.} To determine the variance of each modality, we need to measure the uncertainty of its prediction output. The current estimation of uncertainty via the self-information entropy is computationally efficient yet may be imperfect since it only leverages the information of a single inference. Intuitively, more accurate uncertainty estimation can lead to more accurate modulation coefficients. However, how to balance the complexity and accuracy of the uncertainty estimation is a challenging problem.  We leave this problem to future work.

 % 

% \textbf{Broader Impacts.} In this paper, we re-analyze the under-optimized problem in multimodal learning and reveal that balanced learning is not optimal for multimodal learning. In contrast, the imbalanced learning obeying morality modality can contribute to better performance. This provides an insightful view of studying under-optimized multimodal learning. Besides, the proposed ARL strategy is independent of the structures and fusion methods of the multimodal model. It could be used in various scenarios beyond audio-visual classification. 